\documentclass[10pt,twocolumn,letterpaper]{article}

\usepackage{cvpr}
\usepackage{times}
\usepackage{epsfig}
\usepackage{graphicx}
\usepackage{amsmath}
\usepackage{amssymb}
\usepackage{algorithm}
\usepackage{algpseudocode}
\usepackage{multirow}
\usepackage{epstopdf}
\usepackage{authblk}

\usepackage[pagebackref=true,breaklinks=true,letterpaper=true,colorlinks,bookmarks=false]{hyperref}

\algnewcommand\algorithmicinput{\textbf{INPUT:}}
\algnewcommand\INPUT{\item[\algorithmicinput]}
\algnewcommand\algorithmicoutput{\textbf{OUTPUT:}}
\algnewcommand\OUTPUT{\item[\algorithmicoutput]}

 \cvprfinalcopy 

\def\cvprPaperID{} 

\ifcvprfinal\fi
\begin{document}

\title{A Pursuit of Temporal Accuracy in General Activity Detection}

\author[1]{Yuanjun Xiong}
\author[1]{Yue Zhao}
\author[2]{Limin Wang}
\author[1]{Dahua Lin}
\author[1]{Xiaoou Tang}

\affil[1]{Department of Information Engineering, The Chinese University of Hong Kong}
\affil[2]{Computer Vision Laboratory, ETH Zurich, Switzerland}

\maketitle

\begin{abstract}
Detecting activities in untrimmed videos is an important but challenging task. 
The performance of existing methods remains unsatisfactory, 
\eg~they often meet difficulties in locating the beginning and end of a long complex action. 
In this paper, we propose a generic framework that can 
accurately detect a wide variety of activities from untrimmed videos. 
Our first contribution is  a novel proposal scheme that 
can efficiently generate candidates with accurate temporal boundaries.
The other contribution is a cascaded classification pipeline that explicitly distinguishes between relevance and completeness of a candidate instance.  
On two challenging temporal activity detection datasets, THUMOS14 and ActivityNet, 
the proposed framework significantly outperforms the existing state-of-the-art methods, 
demonstrating superior accuracy and strong adaptivity in handling activities with various temporal structures.
\end{abstract}


\section{Introduction}
\label{sec:intro}

Detecting human activities is crucial to video understanding.
This task has long been an important research topic in computer vision,
and is gaining even more attention in recent years, due to the explosive growth of video data.
Activity detection aims to answer two questions:
(1) \textbf{what} the activity is and (2) \textbf{when} it starts and ends.
Thanks to the advances in deep learning,
the past few years witnessed substantial progress in 
action recognition~\cite{Simonyan14TwoStream,Tran15C3D,WangQT15TDD,Wang2016TSN}.
These methods perform reasonably well in answering the question of \emph{``what''}, 
namely, they can recognize the class of an action with reasonable accuracy from a well-trimmed video.
Nonetheless, recognition methods as such are not capable of answering the other question -- \emph{``when''}.
This is a major obstacle we face when working with \emph{untrimmed videos},
\ie~those in which the actual activities only last for a fraction of the entire duration.

\begin{figure}
\centering
\includegraphics[width=0.9\linewidth]{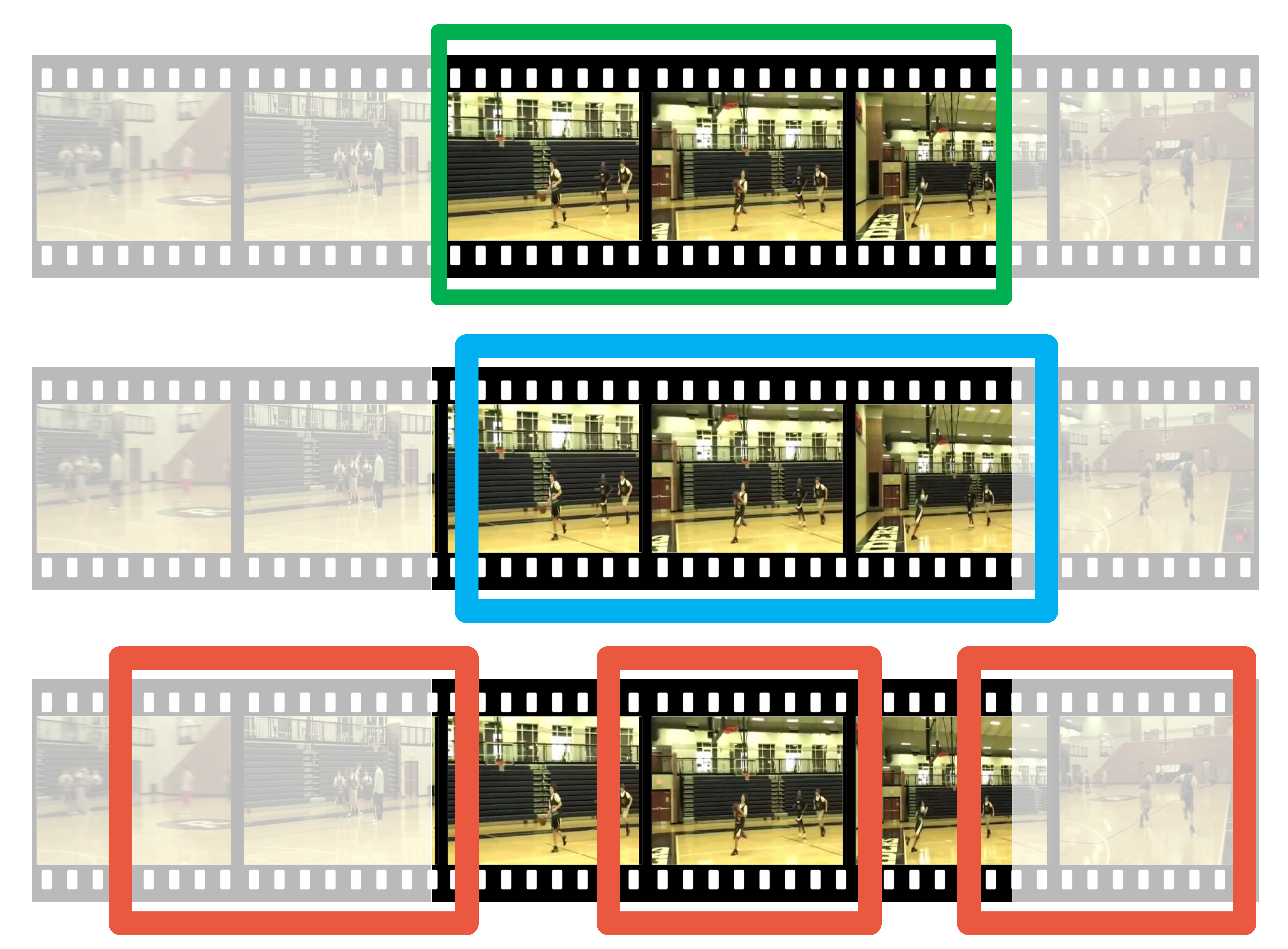}
\caption{An example of temporal action detection. The green box denotes an instance of the activity class ``LayUp Drill'' in the ActivityNet~\cite{caba2015activitynet} datasets.
	The blue box denotes a good detection result. The red boxes demonstrate some cases of bad localization and false detections.
	Note although the red box in the center has less than  $ 0.4 $ IOU with the groundtruth instance, it does include a very representative part of the action instance.}
\label{fig:teaser}
\end{figure}

In the real world, however, \emph{untrimmed videos} dominate. 
From YouTube to Vimeo, from surveillance systems to personal movies, 
most videos are untrimmed. 
This motivates the community to shift its attention to the problem
of detecting activities in untrimmed videos.
Whereas some attempts~\cite{Yeung2016FrameGlimpse,Yuan2016ScorePyramids}
have been made to tackle this problem, 
the performance of existing methods remains far from satisfactory. 
On ActivityNet 2016~\cite{caba2015activitynet}, the top detection method
reports an mAP at $42.5\%$ with $0.5$-IOU threshold.
As the threshold increases to $0.75$ and $0.95$, the mAP dramatically
reduces to $2.88\%$ and $0.06\%$. 
Other methods also see serious performance degradation when the threshold increases.
Clearly, these methods, which are considered to represent
the state of the art, are lacking in the capability of \emph{accurate detection}. 
The difficulties primarily stem from several key challenges
that have yet to be solved. 

First and foremost, it is nontrivial to tell whether a video segment
captures an entire action or just a part of it.  
This issue is particularly prominent for detection methods based on region proposals,
due to a key distinction between \emph{temporal proposals} from \emph{spatial proposals}:
In an image, the appearance of an object often looks quite different 
from a local part of it. Hence, it is generally not very difficult for a visual detector
to tell whether a window corresponds to an entire object or just a local part. 
However, for a video, one can often easily tell what the action is from just a small segment
(or even a single frame), but the entire action may actually last much longer~\cite{Schindler2008Snippet}. 
The obscured distinction between an action and a part thereof makes it very difficult to 
accurately locate the starting and ending points. 

Second, the duration of an action can vary significantly, 
from a second to several minutes. 
This significant variation in length poses two challenges:
(1) Methods relying on sliding windows will meet substantial difficulties 
adapting to actions of different lengths. 
To obtain high localization accuracy, a large number of window scales and 
small sliding steps would be needed, which can lead to dramatically increased
computational cost.
(2) Conventional action recognition methods mostly operate on densely sampled
frames. Hence, processing long actions is very expensive. 
This issue can limit their capability of analyzing long actions and make it 
difficult to perform end-to-end learning.

In this work, we aim to develop a new framework that can 
detect activities of \emph{various lengths} from untrimmed videos,
and \emph{accurately} locate their temporal boundaries.  
This framework adopts the \emph{``proposal + classification''}
paradigm, which has been very successful 
in object detection~\cite{Gu2009RegionIdea,Girshick2014RCNN}.
It is however worth noting that despite its success in 
\emph{spatial detection}, extending it to \emph{temporal detection} 
is nontrivial, due to the challenges discussed above. 

In tackling these challenges, we develop several innovative techniques. 
First, we propose a learning-based bottom-up scheme to generate temporal proposals, 
called \emph{temporal actionness grouping}. 
In contrast to the methods that rely on sliding windows, this scheme makes no assumption on 
the activity durations and thus can work with activities with significantly varying lengths. 
Also, thanks to its bottom-up nature, the resultant proposals tend to be more sensitive to 
temporal boundaries than the sliding windows generated following a regular scheme. 
Second, we propose a cascaded classification pipeline that treats two kinds of false proposals,
namely, \emph{background proposals} and \emph{incomplete proposals}, differently. 
Specifically, it filters out all background proposals in the first stage, 
and then removes incomplete proposals, 
\ie~those corresponding to a sub-segment instead of the entire actions, 
using a \emph{completeness filter}. 
We found empirically that this cascaded approach is very effective in distinguishing 
the complete action proposals from others. 
Moreover, we adopt the sparse snippet sampling method
introduced in~\cite{Wang2016TSN}, which can substantially reduce the computational cost 
for long action proposals. 

We tested the proposed framework on two challenging datasets, namely,
THUMOS14~\cite{Jiang2014THUMOS14} and ActivityNet~\cite{caba2015activitynet}.
On both datasets, our method outperforms the state of the art across different IOU levels. 
The performance gain is especially remarkable under high IOU thresholds, \eg~$0.7$ or $0.9$.
This demonstrated its superior detection accuracy.
Also note that the temporal structures of the actions in ActivityNet are very 
different from those in THUMOS14. 
The consistent high performance on both datasets also shows the method's strong adaptivity.

\section{Related Work}
\label{related}

\paragraph{Action Recognition.}
Action recognition has been extensively studied in the past few years~\cite{Laptev05STIP,WangS13IDT,Simonyan14TwoStream,Tran15C3D,WangQT15TDD,Wang2016TSN,ZhangWW0W16}.
Earlier methods are mostly based on hand-crafted visual features~\cite{Laptev05STIP,WangS13IDT}.
In past several years, deep learning has resulted in great performance gain.
Convolutional Neural Networks (CNNs) are first introduced to this task in~\cite{KarpathyCVPR14Sports1M}.
Later, two-stream architectures~\cite{Simonyan14TwoStream} and 3D-CNN~\cite{Tran15C3D} 
are proposed to incorporate both appearance and motion features.
There have also been efforts that explore the use of long-range temporal structures~\cite{WangQT15TDD,Ng15BeyondSnippet,DonahueJ2015LRCN}.
Recently, Wang \etal~\cite{Wang2016TSN} introduced a segmental architecture
that can efficiently handle longer videos via sparse sampling.
Most action recognition methods assume that the input videos are well-trimmed,
\ie~the action of interest lasts for nearly the entire duration. 
Hence, they usually ignore the localization issue and can focus on classification.

\begin{figure*}[t]
	\centering
	\includegraphics[width=\linewidth]{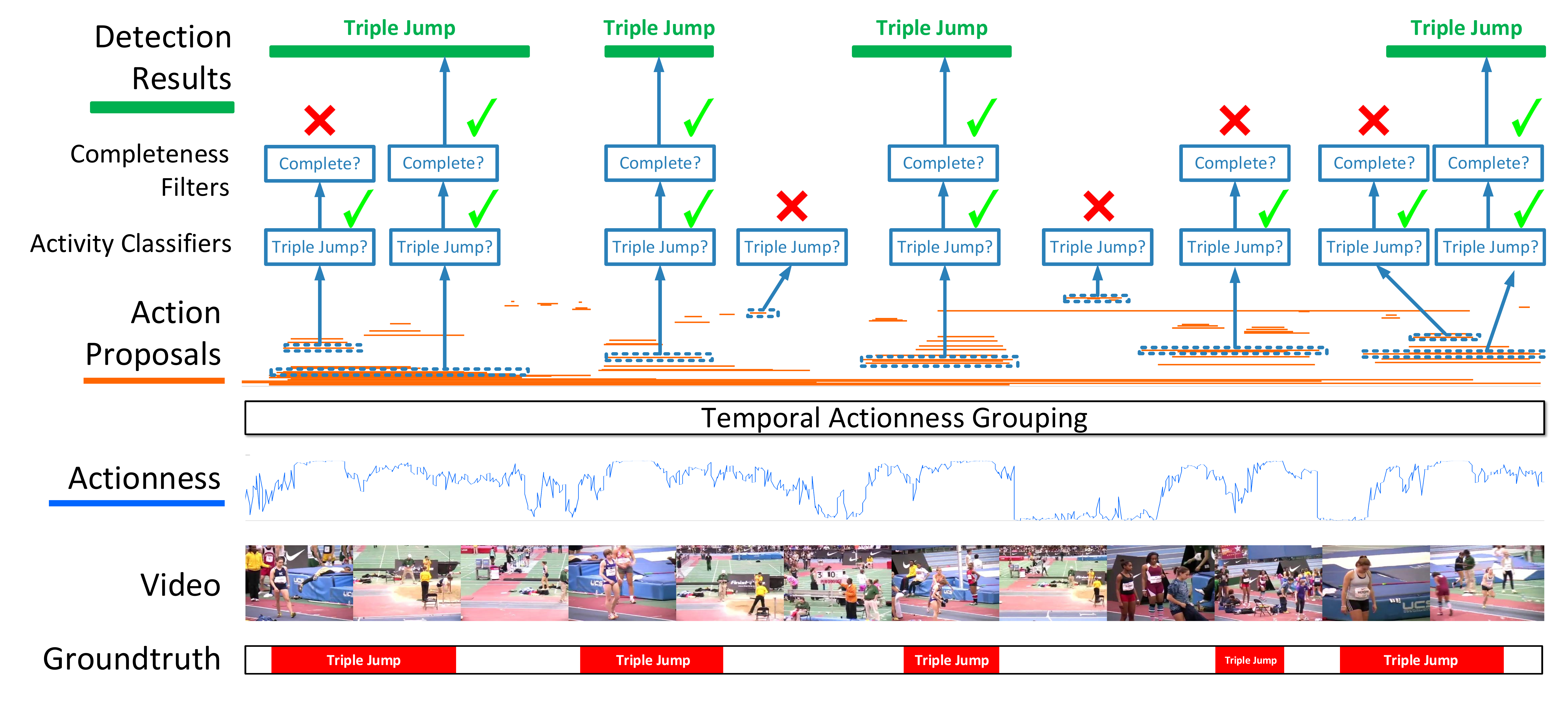}
	\caption{An overview of the proposed framework. This video from the ActivityNet~\cite{caba2015activitynet} dataset contains five instances of ``Triple Jump'' class. The proposed action detection framework starts with evaluating the actionness of the snippets of the video. 
		A set of temporal action proposals (in orange color) are generated with temporal actionness grouping (TAG). The proposals are evaluated against the cascaded classifiers to verify their relevance and completeness. Only proposals being complete instances of triple jumping are produced by the framework. Notice how \emph{non-complete} proposals and background proposals are rejected by the framework.}
	\label{fig:overview}
\end{figure*}

\vspace{-8pt}
\paragraph{Object Detection.}
Recent object detection methods are good examples of 
how we can transfer the knowledge we learned in recognition tasks to the problem of detection.
Mainstream approaches for object detection~\cite{Girshick2014RCNN,Girshick2015FRCNN,Ren2015FasterRCNN} 
usually follow the paradigm of \emph{proposal + classification}.
The proposals are usually generated by bottom-up methods that exploit low-level cues~\cite{Van2011SS,Dollar2014Edgebox}.
The Faster RCNN framework in~\cite{Ren2015FRCNN} uses a neural network to generate proposals,
resulting in improved performance. 
Compared to methods that rely on sliding windows, these methods that utilize visual cues to generate
proposals usually show considerably better performance while requiring less candidates~\cite{Girshick2014RCNN}.

\vspace{-8pt}
\paragraph{Temporal Action Detection.}
Previous works on activity detection mainly use sliding windows as candidates 
and focus on feature representations and 
classification~\cite{Gaidon2013Actom,Tang2013RightFeature,Oneata2013FV,Mettes2015Bofrag,Yuan2016ScorePyramids}.
Recent works incorporate deep networks into the detection frameworks and obtain improved performance~\cite{Yeung2016FrameGlimpse,Shou2016SCNN}.
In~\cite{Yeung2016FrameGlimpse}, 
a Recurrent Neural Network (RNN) is introduced, which takes frame-wise features as inputs
and predicts the starting and ending points of the actions. 
However, in this work, the CNN for feature extraction and the higher-level RNN are trained separately,
which may lead to sub-optimal performance -- as the CNN features may not be suitable for temporal localization.
Empirically, we also found that the predicted temporal boundaries are not very accurate.
In~\cite{Shou2016SCNN}, a \emph{proposal CNN} is used to filter out background windows 
and a \emph{localization CNN} to improve the localization accuracy.
This method relies on C3D~\cite{Tran15C3D}.
As C3D needs to be trained on clips of $16$ frames with no large intervals in between,
this method only handles temporal windows spanning at most $512$ frames (about $17$ seconds). 
This issue severely limits its application in real-world scenarios.
Our method is distinct from these approaches in two aspects.
First,using a novel bottom-up proposal scheme, it can generate more accurate candidates
and can handle a wide range of action lengths.
Second, the cascaded classifier design distinguishes the tasks of measuring the relevance and evaluating the completeness of a candidate instance, which we found rather important to achieve reasonable temporal localization accuracies.

\section{Framework Overview}
\label{sec:overview}

The proposed framework, as shown in Figure~\ref{fig:overview},  
comprises two stages: 
\emph{generating temporal proposals} and
\emph{classifying proposed candidates}. 
The former is to produce a set of class-agnostic temporal regions 
that potentially reflect actions of interest,
while the latter is to determine whether each candidate actually
corresponds to an action and what class it belongs to.
While such a \emph{proposal + classification} paradigm is widely 
adopted in image object detection~\cite{Gu2009RegionIdea,Girshick2014RCNN},
its use in action detection, as discussed earlier, still faces several key challenges.
These challenges include the obscured distinction between \emph{wholes} and \emph{parts} and
the substantial variation in action duration.
As a key contribution of this work, we propose novel designs for each stage 
to tackle these challenges.

Specifically, for temporal proposal generation,
it is desirable to have regions that can cover a wide range of durations
while accurately matching the ground-truths. 
Unlike previous methods that rely on sliding windows~\cite{karaman2014fast,Yuan2016ScorePyramids}, 
we devise a new proposal generation method called \emph{Temporal Actionness Grouping (TAG)},
where a convolutional network is learned to distinguish between \emph{action} 
and \emph{background}, \ie~those snippets that reflect no actions of interest. 
Given a video, we first sample a sequence of snippets, 
then use this network to produce \emph{actionness} scores for them, 
and finally group them into temporal regions of various granularities in a bottom-up manner. 
This method can generate proposals with vastly varying lengths.
Also, owing to the high sensitivity of the bottom-up procedure to temporal transitions, 
the boundaries of the derived proposals are often quite accurate.

With a set of proposed temporal regions, the next stage is to 
classify them into action classes, while filtering out those that 
do not actually capture a complete action. 
As mentioned, distinguishing between 
\emph{complete actions} and incomplete ones is an important challenge
in action detection. 
Note that previous methods have also attempted to tackle this issue.
For example, the method in~\cite{Shou2016SCNN} considers incomplete actions
as background, which often leads to confusion when training the background/action 
classifier. 
We argue that \emph{actionness} and \emph{completeness} are essentially different
characteristics,  and thus design a cascaded classification pipeline with two 
steps.
The first step removes those that belong to the background, 
while the second step, which we refer to as \emph{completeness classification},
is dedicated to identifying those candidates that capture only an incomplete 
part of an action and dropping them from the results.

In what follows, we will elaborate on the detailed designs of both stages in turn.

\section{Temporal Region Proposals}
\label{sec:proposal}

The \emph{temporal region proposals} are generated with a bottom-up procedure,
which consists of three steps: 
extract snippets, evaluate snippet-wise \emph{actionness}, and 
finally group them into region proposals. 
Here, each snippet combines a video frame together with an optical flow field
derived therefrom, which conveys not only the scene appearance at a particular time step, 
but also the motion information at the moment. 
Given a video, a sequence of snippets will be extracted with a regular interval in between. 

\emph{Actionness}, as its name suggests, is a class-agnostic measure of the possibility of a snippet residing in any activity instance.
Therefore, activity instances are likely to be found in the parts of a video containing snippets with relative higher actionness.
To evaluate the \emph{actionness}, we learn a binary classifier based on 
the Temporal Segment Network proposed in~\cite{Wang2016TSN}.
It uses whole videos to train two stream CNNs in order to model the long-range temporal dynamics.
In our practice, we modify it to take regions inside videos as input.
To train this classifier, we treat all annotated action instances
as positive regions, and randomly sample negative regions from the part of videos that have no action annotated with the ratio of $ 1:1 $. 

With a sequence of snippets extracted from a video, 
we use the classifier learned as above to evaluate the \emph{actionness} score
for each snippet. 
The values of the scores range from $0$ to $1$, and thus can be interpreted
as the probability of a snippet being in an action.
To generate temporal region proposals, our basic idea is to 
group consecutive snippets with high actionness scores. 
As our goal is to develop a generic scheme that works with a wide range of 
situations, the \emph{robustness to noise} and the capability of handling 
substantially \emph{varying durations} are two desiderata.

\begin{figure}
\centering
\includegraphics[width=\linewidth]{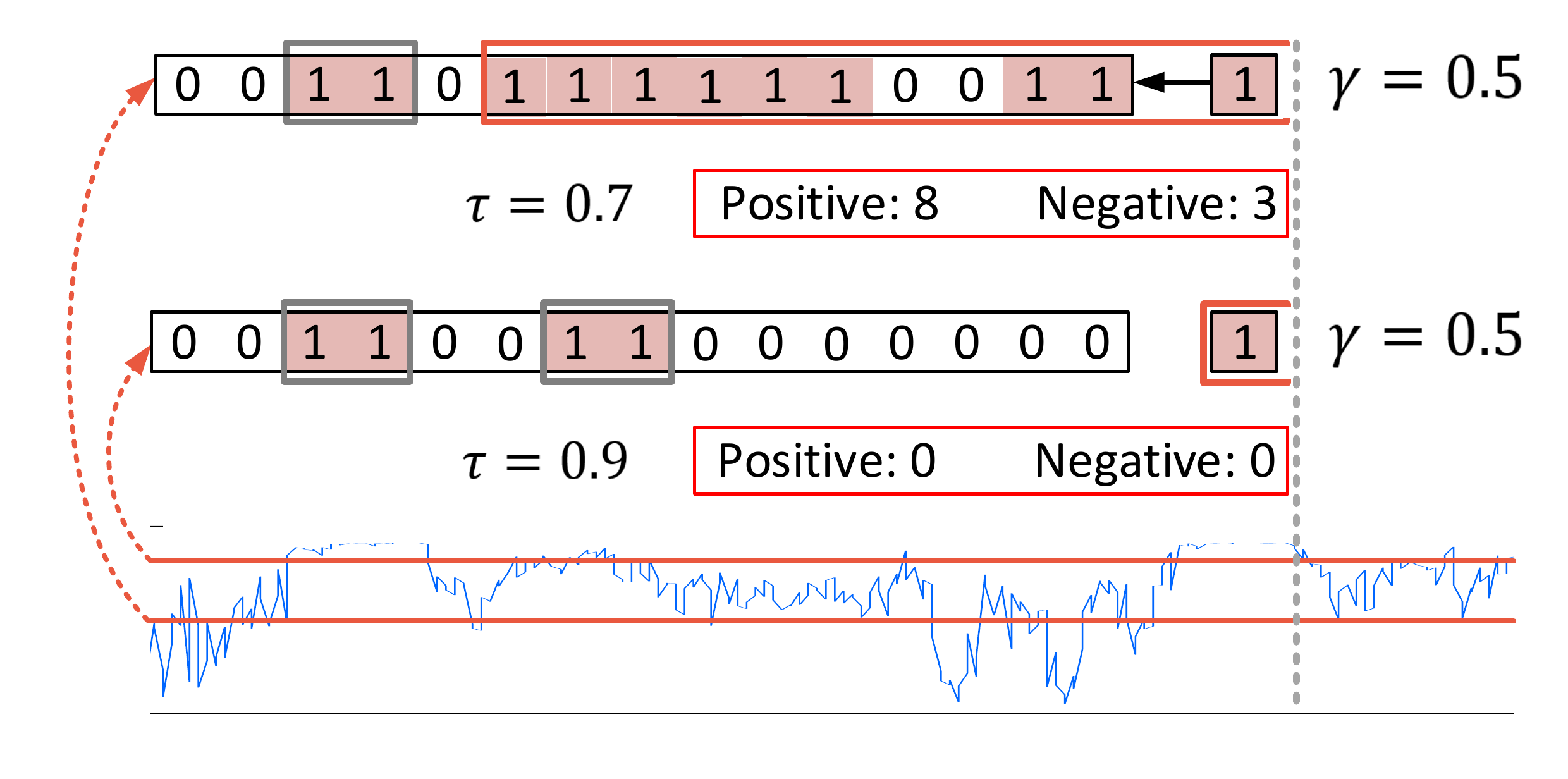}
\caption{An illustration of the temporal actionness grouping algorithm.
	We show two concurrent grouping processes with the foreground/background thresholds $ \tau$ of $ 0.7 $ and $ 0.9 $.
	Each snippet is labeled as ``1'' for foreground or ``0'' for background. The red open box denotes the proposal undergoing grouping. The gray ones are emitted temporal action proposals.
	Note there are several ``0'' snippet grouped because we have not hit the tolerance threshold $ \gamma $.}
\label{fig:tag}
\end{figure}

With these objectives in mind, 
we devise a robust grouping scheme that tolerates occasional outliers,
\eg~a small fraction of low-actionness snippets within an action segment
should be allowed. 
As illustrated in Figure~\ref{fig:tag}, the scheme first obtains a number of 
\emph{action fragments} by thresholding -- a \emph{fragment} here is a
consecutive sub-sequence of snippets whose actionness scores are above
a certain threshold $\tau$. 
Then, to generate a region proposal, we pick a fragment as a starting point
and expand it recursively by absorbing succeeding fragments.
The expansion terminates when the portion of low-actionness snippets goes beyond 
$\gamma$, a positive value which we refer to as the \emph{tolerance threshold}.
Beginning with different fragments, we can obtain a collection of 
different region proposals.

Note that this scheme is controlled by two design parameters:
the \emph{actionness threshold} $\tau$ and the \emph{tolerance threshold} $\gamma$.
It can be seen that different combination of these thresholds will lead to proposals of 
different granularities.
So we use two sets of evenly distributed values for both $\tau$ and $ \gamma $.
The final proposal set is the union of those derived from individual combination of the two values.
This multi-threshold design enable us to generate proposals with diverse granularities.
More importantly, it removes the need for manual parameter tunning on a specific dataset, which is time costing and can not generalize.
Near duplicate proposals collected with this scheme will be pruned using non-maximal suppression 
with an IOU threshold $0.95$.
For a typical video, it would result in a collection that comprises about 
$30$ proposals per minute. 
While the size of the collection is moderate, the proposals therein, however, can often 
cover a very wide range of durations, 
from less than $10 $ frames to more than $ 5000 $ frames. 

We call the proposal generation scheme described above 
\emph{Temporal Actionness Grouping (TAG)}.
Compared to existing methods, it has several advantages:
(1) Thanks to the actionness classifier, the generated proposals are mostly 
focused on action-related contents, which greatly reduce the number of needed proposals.
(2) Action fragments are sensitive to temporal transitions. 
Hence, as a bottom-up method that relies on merging action fragments, 
it often yields proposals with more accurate temporal boundaries. 
(3) With the multi-threshold design, it can cover a broad range of actions
without the need of case-specific parameter tuning.
With these properties, the proposed method can achieve high recall with 
just a moderate number of proposals. 
This also benefits the training of the classifiers in the next stage.

\section{Detecting Action Instances}
\label{sec:detector}

With a set of candidate temporal regions, 
the next stage is to identify \emph{complete} action instances therefrom and 
classify them to specific action categories.
As mentioned, this is accomplished by a cascaded pipeline with two steps:
\emph{activity classification} and \emph{completeness filtering}.
The first step removes those belonging to the background and 
classifies the remaining ones. 
The retained subset may still contain \emph{incomplete} or
\emph{over-complete} instances.
The second step will filter out such proposals, using 
class-specific \emph{completeness filters}. 
With task of completeness testing separated from 
activity classification, we no longer need to entangle two essentially different goals,
which, as we empirically found, could severely confuse the classifiers.

\subsection{Activity Classification}
\label{sec:detector/cls}

We base the activity classifiers on TSN~\cite{Wang2016TSN}.
During training, region proposals that overlap with a ground-truth instance
with an IOU above $0.7$ will be used as positive samples. 
In selecting negative samples, a different criterion is used.
Instead of using IOU, we consider a proposal as a negative sample
only when less than $5\%$ of its time span overlaps with any annotated instances.
The rationale behind is that incorrectly localized samples, \eg~those only cover
small parts of an action, can also have low IOU values. 
However, if we consider them as negative samples, the activity classifier
can be severely confused, as they can still contain parts of action that are very discriminative~\cite{Schindler2008Snippet}.
Using the modified criterion above, we can preclude such samples
from being fed to the training set, such that
the activity classifiers can focus on distinguishing between the actions of interest
and the background. 

In the testing, the learned classifiers will be applied to a video at a fixed frame rate, producing the classification scores for each sampled snippet. 
For each region proposal, the snippet-wise classification scores are aggregated into
region-level scores to classify a proposal to its activity class or background.
Only the proposals classified as non-background classes will be retained for completeness filtering.

\begin{figure}
\centering
\includegraphics[width=\linewidth]{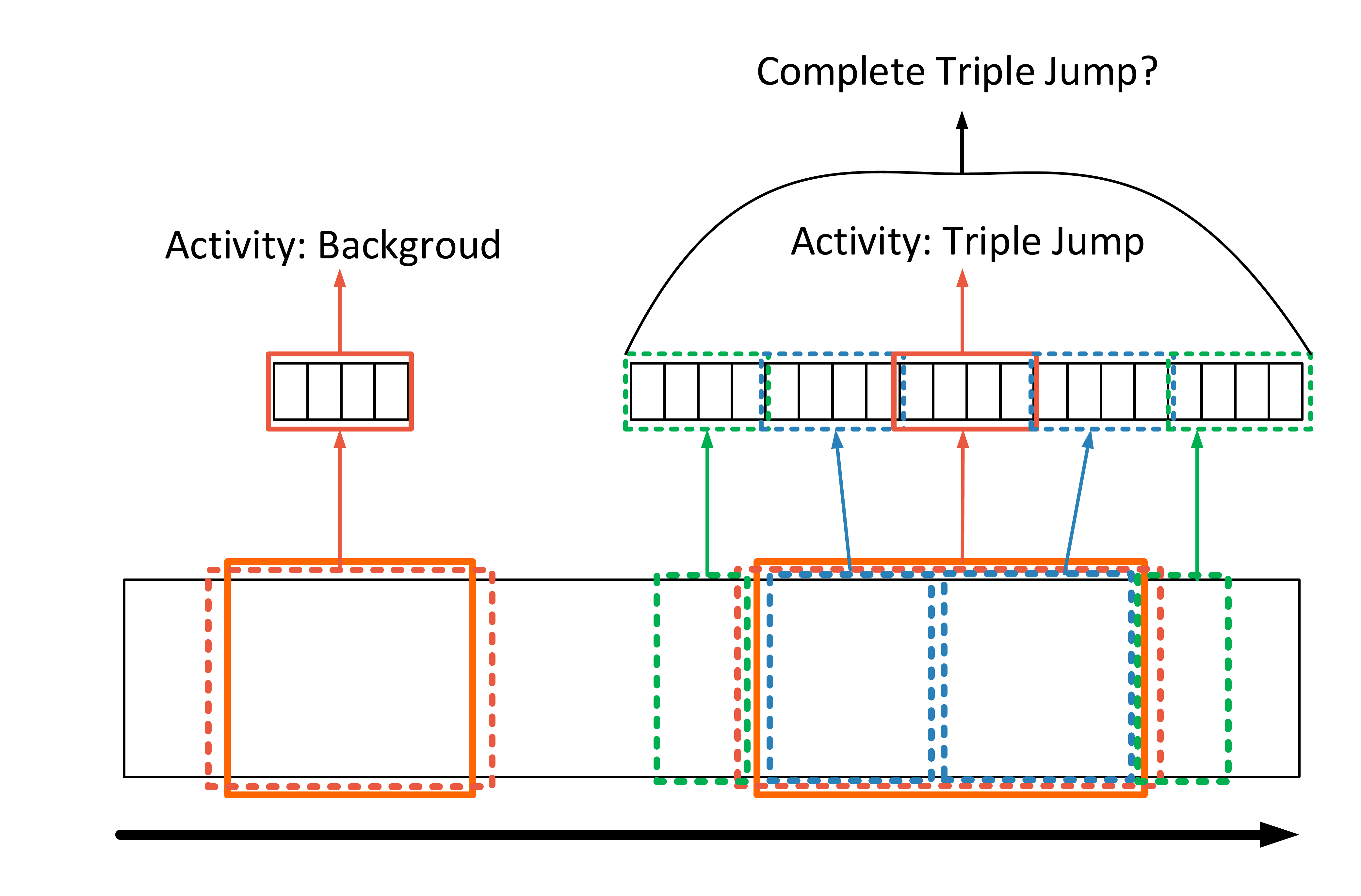}
\caption{The proposal classification module. The activity classifiers first remove background proposals and classify the proposals to its activity class.
	Then the class-aware completeness filters evaluate the remaining proposals using features from the temporal pyramid and surrounding fragments.}
\label{fig:detector}
\end{figure}

\subsection{Completeness Filtering}
\label{sec:detector/loc}

The second step is to identify those incorrectly localized candidates from the remaining ones,
including both \emph{incomplete} and \emph{over-complete} action instances. 
The key question here is {\em what characteristics are good indicators of completeness}.

We observe that it is often quite difficult to tell whether a video segment covers an entire action or not
by only looking at the individual snippets inside it. 
The judgment would become much more obvious if we also look at: 1) the differences between different parts of the segment and 2) the surroundings, \eg~what happens
before and after the segment.
Inspired by this, we devise a simple feature representation that reuses the activity classification scores, 
which comprises three parts:
(1) A temporal pyramid of two levels. The first level pools the snippet scores within the proposed region. The second level split the segment into two parts and pool the snippet scores inside each part.
(2) The average classification scores of two short periods -- the ones before and after the proposed region.
The combination of these features characterizes not only the temporal structures inside but also 
the contexts that come before and after. 
Since the temporal structures of different kinds of activities vary significantly,
we train class-specific SVMs on such a representation, one for each class, with hard negative mining.

Combining both steps, we can obtain the final \emph{detection confidence} for each proposal, as 
\begin{equation}
	S_{Det} = P_{a} \times \exp(S_c),
\end{equation}
where $P_a$ is the probability from the activity classifier and $S_c$ is the completeness score.
This formulation denotes that the final confidence of a detection result is the combination of its class probability and completeness score.
Non-maximal suppression is employed to remove duplicate detections with IOU thresholds of $0.6$ on ActivityNet datasets and $ 0.2 $ on THUMOS14 following~\cite{Shou2016SCNN}.
Our experiments showed that the completeness filters can effectively remove 
incorrectly localized proposals.
Also, due to the simple feature design, the cost of the completeness filtering is insignificant.

\section{Experimental Results}
\label{sec:experiment}

In this section, we evaluate the effectiveness of the proposed framework on standard benchmarks.
We first introduce the evaluation datasets and the implementation details of our method, 
and then explore the effect of components in the proposed framework.
Finally, we compare the performance of the proposed framework with other state-of-the-art approaches.

\subsection{Datasets}

In this work, we conduct experiments on two large-scale benchmark datasets:
\textbf{ActivityNet}~\cite{caba2015activitynet} and
\textbf{THUMOS14}~\cite{Jiang2014THUMOS14}.
ActivityNet~\cite{caba2015activitynet} datasets have two versions, \emph{v1.2} and \emph{v1.3}.
The former contains $9682$ videos in $100$ classes, while the latter, which is a superset of v1.2 and used in the ActivityNet Challenge 2016, contains $19,994$ videos in $200$ classes.
Videos in each version of the dataset are divided into three disjoint subsets,
training, validation, and testing, with a $2:1:1$ split.
%

The THUMOS14~\cite{Jiang2014THUMOS14} dataset has $1010$ videos for validation and $1574$ videos for testing.
These videos contain annotated action instances that belong to $20$ activity classes. 
This dataset does not provide the training set by itself.
Instead, the UCF101~\cite{Soomro2012Ucf101} is appointed as the official training set.
As no temporal annotations are provided by this training set, 
we train the detectors on the validation set and evaluate the performance 
on the testing set.

\subsection{Implementation Details}

We use SGD to learn CNN parameters in our framework, with the batch size of $ 128 $ and momentum as $ 0.9 $.
The network training is conducted with the publicly available TSN toolbox~\cite{Wang2016TSN} and Caffe~\cite{Jia2014Caffe}.
We initialize the CNNs with pre-trained models from ImageNet~\cite{Deng2009ImageNet}.
The initial learning rates are set to $ 0.001 $ for RGB networks and $ 0.005 $ for optical flow networks.

We use an actionness classifier trained on ActivityNet v1.2 on all three datasets for proposal generation. Its RGB CNN is trained for $ 2500 $ iterations on ActivityNet v1.2's training set and $ 18000 $ iterations for its optical flow CNN.
For detectors, the RGB CNN is trained for $9500 $ iterations and $ 20000 $ for optical flow CNN on both versions of ActivityNet.
The learning rates are decreased after every $ 4000 $ and $ 9000 $ iterations, respectively.
For THUMOS14, because there are only about $ 220 $ videos related to the $ 20 $ action classes in the training set, 
we train RGB CNN for $ 1000 $ iterations and optical flow CNN for $ 6000 $ iterations. 
The learning rates are decreased after every $ 400 $ and $ 2500 $ iterations, respectively.
The models and the source codes will be released
\footnote{https://github.com/yjxiong/action-detection}.

\subsection{Evaluation Metrics}\label{sec:exp/metrics}

As both datasets come from action recognition challenges, 
each dataset has its own convention of reporting performance metrics.
We follow their conventions, reporting mean average precision (mAP) at different IOU thresholds.
To obtain decent mAP values at high IOU criteria, the detected instances must have correct class labels and accurately locate starting and ending time of the complete instances. 
So these mAP values are also good measures for the accuracy of temporal localization.
On both versions of ActivityNet, the IOU thresholds are $ \{0.5, 0.75, 0.95\} $. 
The average of mAP values with IOU thresholds $ [0.5:0.05:0.95] $ are also reported on ActivityNet.
On THUMOS14, the IOU thresholds are $ \{0.1, 0.2, 0.3, 0.4, 0.5\} $.

\subsection{Exploration Study}

We first conduct experiments to evaluate the effectiveness of individual components in the proposed framework and investigate how they contribute to 
performance improvement.


\begin{table}[h]
	\begin{center}
		\begin{tabular}{l|c|c|c|c}
			\hline 
			\multirow{2}{*}{Proposal Method}& \multicolumn{2}{|c|}{\textbf{THUMOS14}}  & \multicolumn{2}{|c}{\textbf{ActivityNet v1.2}} \\ 
			\cline{2-5}
			& \# Prop. & AR & \# Prop. & AR \\
			\hline 
			Sliding Windows&  204  & 21.2 & 100 & 34.8\\  
			\hline
			SCNN-prop~\cite{Shou2016SCNN} & 200 & 20.0 & - & - \\
			\hline
			TAP~\cite{caba2016cvpr}& 200  & 23.0  & 90  & 14.9 \\  
			\hline
			DAP~\cite{Escorcia2016DAP}& 200 & 37.0 &100 & 12.0  \\ 
			\hline\hline
			TAG & 117 & 36.9 & 56 & 66.7 \\ 
			\hline 
		\end{tabular} 
	\end{center}
	\caption{Comparison between different temporal action proposal methods. ``AR'' refers to the average recall rates. ``-'' refers to the case where the result is not available.}
	\label{table:proposals}
\end{table}

\begin{table}[h]
	\small
	\begin{center}
\begin{tabular}{c|c|c|c|c}
	\hline
	& \multicolumn{2}{c|}{\textbf{THUMOS14}} & \multicolumn{2}{c}{\textbf{ActivityNet v1.3}} \\ \cline{2-5}
	& Overlapped    & Unseen & Overlapped       & Unseen    \\
	& (10 classes)  & (10 classes)   & (100 classes)    & (100 classes)     \\ \hline
	AR & 46.6          & 28.3           & 68.1             & 66.4              \\ \hline
\end{tabular}
\end{center}
\caption{Study on the ability of the proposal scheme to work with unseen activity classes. We show the average recalls on the two dataset w.r.t. seen and unseen classes for the underlying actionness classifier trained on ActivityNet v1.2}
\label{table:actionness}
\end{table}

\begin{table}[h]
	\small
	\begin{center}
		\begin{tabular}{l|c|c} 
			\hline
			\multirow{2}{*}{Proposal Method} & \multicolumn{2}{c}{\textbf{mAP@0.5}} \\ \cline{2-3} 
			& \textbf{THUMOS14}         & \textbf{ActivityNet v1.2} \\ \hline
			\multirow{2}{*}{Sliding Windows}  &   29.82  &   35.33      \\
			& (787 prop., 62.3AR) & (486 prop., 70.9AR)
			\\ \hline
			\hline
			TAG                                     &   28.25   &   41.13         \\ \hline
		\end{tabular}
	\end{center}
	
	\caption{Detection performance of using different action proposals, measured by mAP. (activity classifier is TSN+Inception-v3; completeness filtering is included).}
	\label{table:proposal_det}
\end{table}

\paragraph{Candidate Region Proposal Methods}
It is known that in image-based object detection, sparse object candidates are better for detector training and improving detection results. 
But {\em is it also true for the temporal action detection task?}
To answer this, we compare the performance of different candidate proposing methods,
using the average recall with IOU thresholds from $ 0.5 $ to $ 0.95 $ as the performance metrics.

A representative for the dense proposal methods is the sliding window approach. 
We generate windows in $ 20 $ exponential scales starting from $ 0.3 $ second long, and slide them through the whole video with a step size of $ 0.4 $ times of window length.
This setting allow the sliding window approach to generate roughly two times the number of proposal in average compared with TAG.

In comparison, we also include other state-of-the-art sparse proposal methods, including TAP~\cite{caba2016cvpr}, DAP~\cite{Escorcia2016DAP}, and the proposal networks in~\cite{Shou2016SCNN}.
For these compared sparse proposal methods, we also have them to produce twice the average number of proposals compared with TAG.
The only exception is TAP~\cite{caba2016cvpr} on ActivityNet v1.2 where the publicly available proposal list has on average $ 90 $ proposals per video.
Results of average recall rates are shown in Table~\ref{table:proposals}. 
These results show that
compared to other methods, TAG can achieve significantly higher recall with considerably less proposals.

Another desirable property of TAG is that it is based on a class-agnostic actionness classifier. This may enable it to work on unseen activity classes. 
To investigate this conjecture, we use the actionness classifier trained on ActivityNet v1.2 to generate proposals on THUMOS14 and ActivityNet v1.3.
In table~\ref{table:actionness}, it is observed that from overlapped classes with ActivityNet v1.2 to those unseen classes, there is no severe performance drop.
This clearly demonstrates the generalization capacity of our proposal scheme.

%

Additionally, we evaluate the effectiveness of using sparse proposals for temporal action detection.
We compare the detection performance of using TAG and sliding windows for proposals.
Here we adjust sliding window configurations to achieve higher AR with much more proposals.
The results shown in Table~\ref{table:proposal_det} suggest that using sparse proposals from TAG, we can achieve a comparable mAP on THUMOS14 and a much better mAP on ActivityNet v1.2.

\paragraph{Activity Classifiers}
It is known that using deeper CNN architectures will benefit action recognition systems~\cite{Wang2016TSN}.
To investigate whether this is also true in temporal action detection, 
we try two architectures, BN-Inception~\cite{IoffeS15BN} and a deeper architecture Inception V3~\cite{Szegedy2016InceptionV3}.
Related results are shown in Table~\ref{table:classifier},
which show that deeper CNN architectures also benefit the task of temporal action detection.

\begin{table}[t]
	\begin{center}
		\begin{tabular}[\linewidth]{l|c}
			\hline
			Activity Classifiers       & mAP@0.5 \\ \hline
			BN-Inception~\cite{IoffeS15BN}                           &    40.29   \\ \hline
			Inception V3~\cite{Szegedy2016InceptionV3}     &    41.13   \\ \hline
		\end{tabular}
	\end{center}
	\caption{Using different CNN architectures for the activity classifiers in the proposed framework. The results are reported on the validation set of ActivityNet v1.2.}
	\label{table:classifier}
\end{table}

\paragraph{Completeness Filters}
In the proposed framework, the detection of activity instances from candidate proposals is performed in two cascaded stages, namely activity classification and completeness filtering.
To investigate the validity of this cascaded design, we perform an ablation study,
whose results are shown in Table~\ref{table:completeness}.

The first baseline is a classification module without the cascade design, which has only one set of classifiers.
The module is trained with foreground/background IOU thresholds of $ 0.7 $ and $ 0.3 $ as in~\cite{Shou2016SCNN}.
This basically means that both background and non-complete region proposals are treated as negative samples.
The single stage classifiers have to distinguish both kinds of false proposals from complete action instances.
This is referred to as ``One Stage'' in Table~\ref{table:completeness}.

Another baseline is replacing the learning-based completeness filters in the cascade with heuristics.
In previous methods~\cite{Shou2016SCNN,Oneata2013FV,wang2014action}, duration-based heuristics are widely used to counter the effect of incomplete instances.
Here we compare with two representative forms of these heuristics.
The first one, ``H1'' multiplies $ T^\alpha $ on the classification scores, where $ T $ is the relative duration of a segment and $ \alpha $ is set to $ 0.7 $ to obtain competitive results on ActivityNet v1.2.
The second baseline heuristics, ``H2'', multiplies the detection scores with the frequencies of the proposal durations~\cite{Shou2016SCNN}.
We can observe in Table~\ref{table:completeness} that the well tuned heuristics cannot work well on both datasets at the same time.
Our learning-based completeness filters perform consistently on different datasets without tunning and produce comparable or superior detection accuracies.

\begin{table}[t]
	\begin{center}
		\begin{tabular}{l|c|c}
			\hline
			\multirow{2}{*}{Module} & \multicolumn{2}{c}{\textbf{mAP@0.5}}                 \\ \cline{2-3} 
			                                     & \textbf{THUMOS14}              & \textbf{ActivityNet v1.2}     \\ \hline
			One Stage                    & 13.96                                     &  8.97            \\ \hline
			Cascade + H1                      & 8.17  & 42.94 \\ \hline 
			Cascade + H2                      & 22.13                                    &                       9.10 \\ \hline 
			
			Cascade + Comp.                & 28.25                                     &           41.13  \\ \hline
		\end{tabular}
	\end{center}
	\caption{Ablation study on the design of candidate classification module. 
		``One Stage'' refers to a single stage of classifiers. 
		``Cascade + H1'' and ``Cascade + H2'' refer to cascade with different duration based heuristics.
		``Cascade + Comp.'' refers to the proposed framework.}
	\label{table:completeness}
\end{table}

\subsection{Comparison with State of the Arts}
Finally we compare our method with other state-of-the-art temporal action detection methods on THUMOS14~\cite{Jiang2014THUMOS14}, ActivityNet v1.2~\cite{caba2015activitynet} and ActivityNet v1.3~\cite{caba2015activitynet}.
To report performance we use the metrics described in Sec.~\ref{sec:exp/metrics}.
Among all the metrics used here, we would like to highlight the average mAP on ActivityNet datasets.
It is the average of the mAP values at IOU thresholds from $ 0.5 $ to $ 0.95 $, 
which could well reflect the ability to accurately localize the temporal boundaries of action instances.

It is also worth noting that the average action duration in THUMOS14 is $4$ seconds, while those in ActivityNet datasets can be as long as $50$ seconds.
Meanwhile the average video duration in THUMOS14 is about twice than those of the ActivityNet datasets ($ 233 $ v.s. $ 114 $).
This difference in the statistics actually reflects the different nature of these datasets in terms of granularities and temporal structures.
Hence, strong adaptivity is required to perform consistently well on both datasets.

\vspace{-10pt}
\paragraph{THUMOS14}
The results on THUMOS14 dataset are shown in Table~\ref{table: thumos14}.
We first compare with the challenge results~\cite{wang2014action,oneata2014lear,Richard2016Language}. 
as well as recent works, including methods using segment-based 3D CNN~\cite{Shou2016SCNN}, 
score pyramids~\cite{Yuan2016ScorePyramids}, and recurrent reinforcement learning ~\cite{Yeung2016FrameGlimpse}.
Using the proposed framework, we are able to outperform previous state-of-the-art methods by over $ 10\% $ in most cases.

\begin{table}[t]
\begin{center}
\begin{tabular}{c|ccccc}
\hline
\multicolumn{6}{c}{\textbf{THUMOS14}, \textbf{mAP@$\alpha$}}                 \\ \hline
Method& 0.1  & 0.2  & 0.3  & 0.4  & 0.5  \\ \hline
Wang \emph{et. al.}~\cite{wang2014action} & 18.2 & 17.0 & 14.0 & 11.7 & 8.3 \\ \hline
Oneata \emph{et. al.}~\cite{oneata2014lear} & 36.6 & 33.6 & 27.0 &  20.8 & 14.4 \\ \hline
Richard \emph{et. al.}~\cite{Richard2016Language} & 39.7 & 35.7 & 30.0 & 23.2 & 15.2 \\ \hline\hline
S-CNN~\cite{Shou2016SCNN} & 47.7 & 43.5 & 36.3 & 28.7 & 19.0 \\ \hline
Yeung \emph{et. al.}~\cite{Yeung2016FrameGlimpse} & 48.9 & 44.0 & 36.0 & 26.4 & 17.1 \\ \hline
Yuan \emph{et. al.}~\cite{Yuan2016ScorePyramids} & 51.4 & 42.6& 33.6&26.1&18.8 \\\hline\hline
Ours & \textbf{64.1} & \textbf{57.7} & \textbf{48.7} & \textbf{39.8} & \textbf{28.2} \\
\hline
\end{tabular}
\end{center}
\caption{Action detection results on THUMOS’14, measured by mean average precision (mAP) for different IoU thresholds $\alpha$. The upper half of the table show challenge results back in 2014.}
\label{table: thumos14}
\end{table}

\vspace{-10pt}
\paragraph{ActivityNet}
The results of ActivityNet v1.2 and v1.3 are shown in Table~\ref{table:anet_v1.2} and Table~\ref{table:anet_v1.3} respectively. 
We first report results on both validation sets.
For references, we list the performance of highest ranking entries in the ActivityNet 2016 challenge.
Finally, we submit our results to the test server of ActivityNet v1.3 and report the detection performance on the testing set.
The proposed framework, using a single model instead of an ensemble, is able to obtain detection performance very close to the winning submission at the IOU threshold of $ 0.5 $.
With a higher IOU threshold, $0.75$ and $0.95$, the detectors are put to test in accurately locating action boundaries.
Under such challenging settings, the proposed detector performs dramatically better than all rivals,
\eg~with IOU threshold $0.75$, we obtain an mAP at $26\%$ (vs $17.8\%$);
with IOU threshold $0.95$, we obtain an mAP at $6.66\%$ (vs $2.88\%$).
In terms of average mAP, we are able to outperform other state-of-the-arts by around $10 \%$.
This clearly demonstrates the superior capability of our method in accurate action detection.

\begin{table}[t]
	\begin{center}
		\begin{tabular}{c|ccc|c}
			\hline
			\multicolumn{5}{c}{\textbf{ActivityNet v1.2} (validation),  \textbf{mAP@$\alpha$}}   \\ \hline
			Method & 0.5 & 0.75 & 0.95 &Average\\ \hline
			Ours-B &    38.67     &     22.94     &     5.23     &       23.58      \\ \hline
			Ours &    41.13     &     24.06     &     5.04     &       24.88      \\ \hline
		\end{tabular}

	\end{center}
	\caption{Action detection results on ActivityNet v1.2, measured by mean average precision (mAP) for different IoU thresholds $\alpha$ and the average mAP of IOU thresholds from $ 0.5 $ to $ 0.95 $..
		In the table ``Ours-B'' refer to the results of using BN-Inception architecture instead of Inception V3.}
	\label{table:anet_v1.2}
\end{table}

\begin{table}[t]
	\begin{center}
		\begin{tabular}{c|ccc|c}
			\hline
			\multicolumn{5}{c}{\textbf{ActivityNet v1.3} (validation), \textbf{mAP@$\alpha$}}   \\ \hline
			Method & 0.5 & 0.75 & 0.95 &Average\\ \hline
			Montes \emph{et. al.}~\cite{Montes_2016_NIPSWS} & 22.51 & - & - &  -      \\ \hline
			Sigh \emph{et. al.}~\cite{DBLP:journals/corr/SinghC16} & 34.47 & - & - &    -         \\ \hline
			Wang \emph{et. al.}~\cite{UTS} & 43.65 &  -  & -  &  -      \\ \hline\hline
			Ours-B& 37.17 & 22.13 & 4.95 & 22.67       \\ \hline
			Ours & 39.12 & 23.48 & 5.49 & 23.98       \\ \hline\hline
			\multicolumn{5}{c}{\textbf{ActivityNet v1.3} (testing),  \textbf{mAP@$\alpha$}}   \\ \hline
			Method & 0.5 & 0.75 & 0.95 &Average\\ \hline
			Wang \emph{et. al.}~\cite{UTS}  & 42.478 & 2.88 & 0.06 & 14.62 \\ \hline
			Singh \emph{et. al.}~\cite{Singh_2016_CVPR} & 28.667 & 17.78 & 2.88 & 17.68 \\ \hline
			Singh \emph{et. al.}~\cite{DBLP:journals/corr/SinghC16}& 36.398 & 11.05 & 0.14 & 17.83 \\ \hline\hline
			Ours& 40.689 & 26.017 & 6.667 &   {\bf 26.05}     \\ \hline
		\end{tabular}
	\end{center}
	\caption{Action detection results on ActivityNet v1.3, measured by mean average precision (mAP) for different IoU thresholds $\alpha$ and the average mAP of IOU thresholds from $ 0.5 $ to $ 0.95 $. In the table, ``Ours-B'' refers to the results using the BN-Inception architecture.}
	\label{table:anet_v1.3}
\end{table}

\subsection{Discussion}
From the experimental results presented above,
we have the following observations:
1) Based on class-agnostic actionness, our TAG proposal method excels in generating temporal proposals and can generalize well to unseen activities.
The sparse proposals generated by TAG are conducive to the detection performance. 
2) The two-stage cascaded design of our classification module is crucial for action detection with high temporal accuracy. It is also a generic design that adapts well to activities with varying temporal structures, such as in THUMOS14 and ActivityNet.

\section{Conclusion}
\label{sec:conclusion}

In this paper, we proposed a generic framework for the task of temporal action detection.
It is built on the proposal + classification paradigm.
By introducing the temporal actionness grouping for action proposal and a cascaded design of proposal classifiers, we achieved significant performance improvement over previous state-of-the-art methods.
Moreover, we demonstrated that our method is both accurate and generic, 
being able to localize temporal boundaries precisely and working well on datasets with different temporal structure of activities.

{\small
\bibliographystyle{ieee}
\bibliography{action_detection}

\begin{thebibliography}{10}\itemsep=-1pt

\bibitem{Deng2009ImageNet}
J.~Deng, W.~Dong, R.~Socher, L.~Li, K.~Li, and F.~Li.
\newblock {ImageNet}: {A} large-scale hierarchical image database.
\newblock In {\em CVPR}, pages 248--255, 2009.

\bibitem{DonahueJ2015LRCN}
J.~Donahue, L.~Anne~Hendricks, S.~Guadarrama, M.~Rohrbach, S.~Venugopalan,
  K.~Saenko, and T.~Darrell.
\newblock Long-term recurrent convolutional networks for visual recognition and
  description.
\newblock In {\em CVPR}, pages 2625--2634, 2015.

\bibitem{Escorcia2016DAP}
V.~Escorcia, F.~Caba~Heilbron, J.~C. Niebles, and B.~Ghanem.
\newblock Daps: Deep action proposals for action understanding.
\newblock In B.~Leibe, J.~Matas, N.~Sebe, and M.~Welling, editors, {\em ECCV},
  pages 768--784, 2016.

\bibitem{caba2016cvpr}
B.~G. Fabian Caba~Heilbron, Juan Carlos~Niebles.
\newblock Fast temporal activity proposals for efficient detection of human
  actions in untrimmed videos.
\newblock In {\em CVPR}, pages 1914--1923, 2016.

\bibitem{caba2015activitynet}
B.~G. Fabian Caba~Heilbron, Victor~Escorcia and J.~C. Niebles.
\newblock Activitynet: A large-scale video benchmark for human activity
  understanding.
\newblock In {\em CVPR}, pages 961--970, 2015.

\bibitem{Gaidon2013Actom}
A.~Gaidon, Z.~Harchaoui, and C.~Schmid.
\newblock Temporal localization of actions with actoms.
\newblock {\em IEEE TPAMI}, 35(11):2782--2795, 2013.

\bibitem{Girshick2015FRCNN}
R.~Girshick.
\newblock Fast r-cnn.
\newblock In {\em ICCV}, pages 1440--1448, 2015.

\bibitem{Girshick2014RCNN}
R.~Girshick, J.~Donahue, T.~Darrell, and J.~Malik.
\newblock Rich feature hierarchies for accurate object detection and semantic
  segmentation.
\newblock In {\em CVPR}, pages 580--587, 2014.

\bibitem{Gu2009RegionIdea}
C.~Gu, J.~J. Lim, P.~Arbel{\'a}ez, and J.~Malik.
\newblock Recognition using regions.
\newblock In {\em CVPR}, pages 1030--1037, 2009.

\bibitem{IoffeS15BN}
S.~Ioffe and C.~Szegedy.
\newblock Batch normalization: Accelerating deep network training by reducing
  internal covariate shift.
\newblock In {\em ICML}, pages 448--456, 2015.

\bibitem{Jia2014Caffe}
Y.~Jia, E.~Shelhamer, J.~Donahue, S.~Karayev, J.~Long, R.~Girshick,
  S.~Guadarrama, and T.~Darrell.
\newblock Caffe: Convolutional architecture for fast feature embedding.
\newblock {\em arXiv preprint arXiv:1408.5093}, 2014.

\bibitem{Jiang2014THUMOS14}
Y.-G. Jiang, J.~Liu, A.~Roshan~Zamir, G.~Toderici, I.~Laptev, M.~Shah, and
  R.~Sukthankar.
\newblock {THUMOS} challenge: Action recognition with a large number of
  classes.
\newblock \url{http://crcv.ucf.edu/THUMOS14/}, 2014.

\bibitem{karaman2014fast}
S.~Karaman, L.~Seidenari, and A.~Del~Bimbo.
\newblock Fast saliency based pooling of fisher encoded dense trajectories.
\newblock In {\em THUMOS Action Recognition Challenge}, 2014.

\bibitem{KarpathyCVPR14Sports1M}
A.~Karpathy, G.~Toderici, S.~Shetty, T.~Leung, R.~Sukthankar, and L.~Fei-Fei.
\newblock Large-scale video classification with convolutional neural networks.
\newblock In {\em CVPR}, pages 1725--1732, 2014.

\bibitem{Laptev05STIP}
I.~Laptev.
\newblock On space-time interest points.
\newblock {\em IJCV}, 64(2-3):107--123, 2005.

\bibitem{Mettes2015Bofrag}
P.~Mettes, J.~C. van Gemert, S.~Cappallo, T.~Mensink, and C.~G. Snoek.
\newblock Bag-of-fragments: Selecting and encoding video fragments for event
  detection and recounting.
\newblock In {\em ICMR}, pages 427--434, 2015.

\bibitem{Montes_2016_NIPSWS}
A.~Montes, A.~Salvador, S.~Pascual, and X.~Giro-i Nieto.
\newblock Temporal activity detection in untrimmed videos with recurrent neural
  networks.
\newblock In {\em NIPS Workshop}, 2016.

\bibitem{Ng15BeyondSnippet}
J.~Y.-H. Ng, M.~Hausknecht, S.~Vijayanarasimhan, O.~Vinyals, R.~Monga, and
  G.~Toderici.
\newblock Beyond short snippets: Deep networks for video classification.
\newblock In {\em CVPR}, pages 4694--4702, 2015.

\bibitem{Oneata2013FV}
D.~Oneata, J.~Verbeek, and C.~Schmid.
\newblock Action and event recognition with fisher vectors on a compact feature
  set.
\newblock In {\em ICCV}, pages 1817--1824, 2013.

\bibitem{oneata2014lear}
D.~Oneata, J.~Verbeek, and C.~Schmid.
\newblock The lear submission at thumos 2014.
\newblock In {\em THUMOS Action Recognition Challenge}, 2014.

\bibitem{Ren2015FasterRCNN}
S.~Ren, K.~He, R.~Girshick, and J.~Sun.
\newblock Faster r-cnn: Towards real-time object detection with region proposal
  networks.
\newblock In {\em NIPS}, pages 91--99, 2015.

\bibitem{Ren2015FRCNN}
S.~Ren, K.~He, R.~Girshick, and J.~Sun.
\newblock Faster {R-CNN}: Towards real-time object detection with region
  proposal networks.
\newblock In {\em NIPS}, pages 91--99, 2015.

\bibitem{Richard2016Language}
A.~Richard and J.~Gall.
\newblock Temporal action detection using a statistical language model.
\newblock In {\em CVPR}, pages 3131--3140, 2016.

\bibitem{Schindler2008Snippet}
K.~Schindler and L.~Van~Gool.
\newblock Action snippets: How many frames does human action recognition
  require?
\newblock In {\em CVPR}, pages 1--8. IEEE, 2008.

\bibitem{Shou2016SCNN}
Z.~Shou, D.~Wang, and S.-F. Chang.
\newblock Temporal action localization in untrimmed videos via multi-stage
  {CNNs}.
\newblock In {\em CVPR}, pages 1049--1058, 2016.

\bibitem{Simonyan14TwoStream}
K.~Simonyan and A.~Zisserman.
\newblock Two-stream convolutional networks for action recognition in videos.
\newblock In {\em NIPS}, pages 568--576, 2014.

\bibitem{Singh_2016_CVPR}
B.~Singh, T.~K. Marks, M.~Jones, O.~Tuzel, and M.~Shao.
\newblock A multi-stream bi-directional recurrent neural network for
  fine-grained action detection.
\newblock In {\em CVPR}, pages 1961--1970, 2016.

\bibitem{DBLP:journals/corr/SinghC16}
G.~Singh and F.~Cuzzolin.
\newblock Untrimmed video classification for activity detection: submission to
  activitynet challenge.
\newblock {\em CoRR}, abs/1607.01979, 2016.

\bibitem{Soomro2012Ucf101}
K.~Soomro, A.~R. Zamir, and M.~Shah.
\newblock Ucf101: A dataset of 101 human actions classes from videos in the
  wild.
\newblock {\em arXiv preprint arXiv:1212.0402}, 2012.

\bibitem{Szegedy2016InceptionV3}
C.~Szegedy, V.~Vanhoucke, S.~Ioffe, J.~Shlens, and Z.~Wojna.
\newblock Rethinking the inception architecture for computer vision.
\newblock In {\em CVPR}, pages 2818--2826, 2016.

\bibitem{Tang2013RightFeature}
K.~Tang, B.~Yao, L.~Fei-Fei, and D.~Koller.
\newblock Combining the right features for complex event recognition.
\newblock In {\em CVPR}, pages 2696--2703, 2013.

\bibitem{Tran15C3D}
D.~Tran, L.~D. Bourdev, R.~Fergus, L.~Torresani, and M.~Paluri.
\newblock Learning spatiotemporal features with {3D} convolutional networks.
\newblock In {\em ICCV}, pages 4489--4497, 2015.

\bibitem{Van2011SS}
K.~E. Van~de Sande, J.~R. Uijlings, T.~Gevers, and A.~W. Smeulders.
\newblock Segmentation as selective search for object recognition.
\newblock In {\em ICCV}, pages 1879--1886, 2011.

\bibitem{WangS13IDT}
H.~Wang and C.~Schmid.
\newblock Action recognition with improved trajectories.
\newblock In {\em ICCV}, pages 3551--3558, 2013.

\bibitem{wang2014action}
L.~Wang, Y.~Qiao, and X.~Tang.
\newblock Action recognition and detection by combining motion and appearance
  features.
\newblock In {\em THUMOS Action Recognition Challenge}, 2014.

\bibitem{WangQT15TDD}
L.~Wang, Y.~Qiao, and X.~Tang.
\newblock Action recognition with trajectory-pooled deep-convolutional
  descriptors.
\newblock In {\em CVPR}, pages 4305--4314, 2015.

\bibitem{Wang2016TSN}
L.~Wang, Y.~Xiong, Z.~Wang, Y.~Qiao, D.~Lin, X.~Tang, and L.~Van~Gool.
\newblock Temporal segment networks: Towards good practices for deep action
  recognition.
\newblock In {\em ECCV}, pages 20--36, 2016.

\bibitem{UTS}
R.~Wang and D.~Tao.
\newblock {UTS} at activitynet 2016.
\newblock In {\em AcitivityNet Large Scale Activity Recognition Challenge
  2016}, 2016.

\bibitem{Yeung2016FrameGlimpse}
S.~Yeung, O.~Russakovsky, G.~Mori, and L.~Fei-Fei.
\newblock End-to-end learning of action detection from frame glimpses in
  videos.
\newblock In {\em CVPR}, pages 2678--2687, 2016.

\bibitem{Yuan2016ScorePyramids}
J.~Yuan, B.~Ni, X.~Yang, and A.~A. Kassim.
\newblock Temporal action localization with pyramid of score distribution
  features.
\newblock In {\em CVPR}, pages 3093--3102, 2016.

\bibitem{ZhangWW0W16}
B.~Zhang, L.~Wang, Z.~Wang, Y.~Qiao, and H.~Wang.
\newblock Real-time action recognition with enhanced motion vector {CNNs}.
\newblock In {\em CVPR}, pages 2718--2726, 2016.

\bibitem{Dollar2014Edgebox}
C.~L. Zitnick and P.~Doll{\'{a}}r.
\newblock Edge boxes: Locating object proposals from edges.
\newblock In {\em ECCV}, pages 391--405, 2014.

\end{thebibliography}
}

\end{document}